\documentclass[10pt,twocolumn,letterpaper]{article}

\usepackage{cvpr}              

\usepackage{graphicx}
\usepackage{amsmath}
\usepackage{amssymb}
\usepackage{booktabs}

\usepackage{amssymb}
\usepackage{colortbl}
\usepackage{graphicx}
\usepackage{multirow}
\usepackage{color}
\usepackage{bbding}
\usepackage[super]{nth}

\definecolor{mygray}{gray}{.9}
\definecolor{light-gray}{gray}{0.72}

\definecolor{citecolor}{RGB}{34,139,34}
\usepackage[pagebackref,breaklinks,colorlinks,citecolor=citecolor]{hyperref}

\makeatletter\renewcommand\paragraph{\@startsection{paragraph}{4}{\z@}
	{.35em \@plus1ex \@minus.2ex}{-.5em}{\normalfont\normalsize\bfseries}}\makeatother

\usepackage[capitalize]{cleveref}
\crefname{section}{Sec.}{Secs.}
\Crefname{section}{Section}{Sections}
\Crefname{table}{Table}{Tables}
\crefname{table}{Tab.}{Tabs.}

\def\cvprPaperID{} 
\def\confName{CVPR}
\def\confYear{2023}

\begin{document}

\title{Unleashing the Power of Visual Prompting At the Pixel Level}

\author{
Junyang Wu\textsuperscript{*1} ~ 
Xianhang Li\textsuperscript{*2} ~ 
Chen Wei\textsuperscript{3} ~ 
Huiyu Wang\textsuperscript{3} ~ 
Alan Yuille\textsuperscript{3} ~  
Yuyin Zhou\textsuperscript{2} ~
Cihang Xie\textsuperscript{2}\\
\small $^{*}$equal technical contribution \vspace{.5em} \\
\textsuperscript{1}Shanghai Jiao Tong University \qquad \qquad \textsuperscript{2}UC Santa Cruz \qquad \qquad \textsuperscript{3}Johns Hopkins University
}
\maketitle

\begin{abstract}
This paper presents a simple and effective visual prompting method for adapting pre-trained models to downstream recognition tasks. Our method includes two key designs. First, rather than directly adding together the prompt and the image, we treat the prompt as an extra and independent learnable component. We show that the strategy of reconciling the prompt and the image matters, and find that warping the prompt around a properly shrinked image empirically works the best. Second, we re-introduce two ``old tricks'' commonly used in building transferable adversarial examples, \ie,  input diversity and gradient normalization, into visual prompting. These techniques improve optimization and enable the prompt to generalize better.

We provide extensive experimental results to demonstrate the effectiveness of our method. Using a CLIP model, our prompting method sets a new record of 82.5\% average accuracy across 12 popular classification datasets, substantially surpassing the prior art by \textbf{+5.2\%}. It is worth noting that this prompting performance already outperforms linear probing by $\textbf{+2.2\%}$ and can even match fully fine-tuning in certain datasets. In addition, our prompting method shows competitive performance across different data scales and against distribution shifts. 
The code is publicly available at \url{https://github.com/UCSC-VLAA/EVP}. 
\end{abstract}


\section{Introduction}
Deep learning models have witnessed pre-training on increasingly large-scale datasets as a general and effective pathway to succeed in both computer vision \cite{MAE, CLIP, bao2021beit} and natural language processing \cite{BERT,gpt3,liu2019roberta}. These pre-trained models are termed foundation models \cite{bommasani2021opportunities}. While fully fine-tuning stands as one of the most prevalent paradigms to effectively adapt these foundation models to a range of downstream tasks, it can be computationally intensive due to the large number of training parameters. This has led to a need for more efficient alternatives for adapting these (cumbersome) foundation models to new tasks.

\begin{figure}[t!]
\centering
\includegraphics[width=\linewidth]{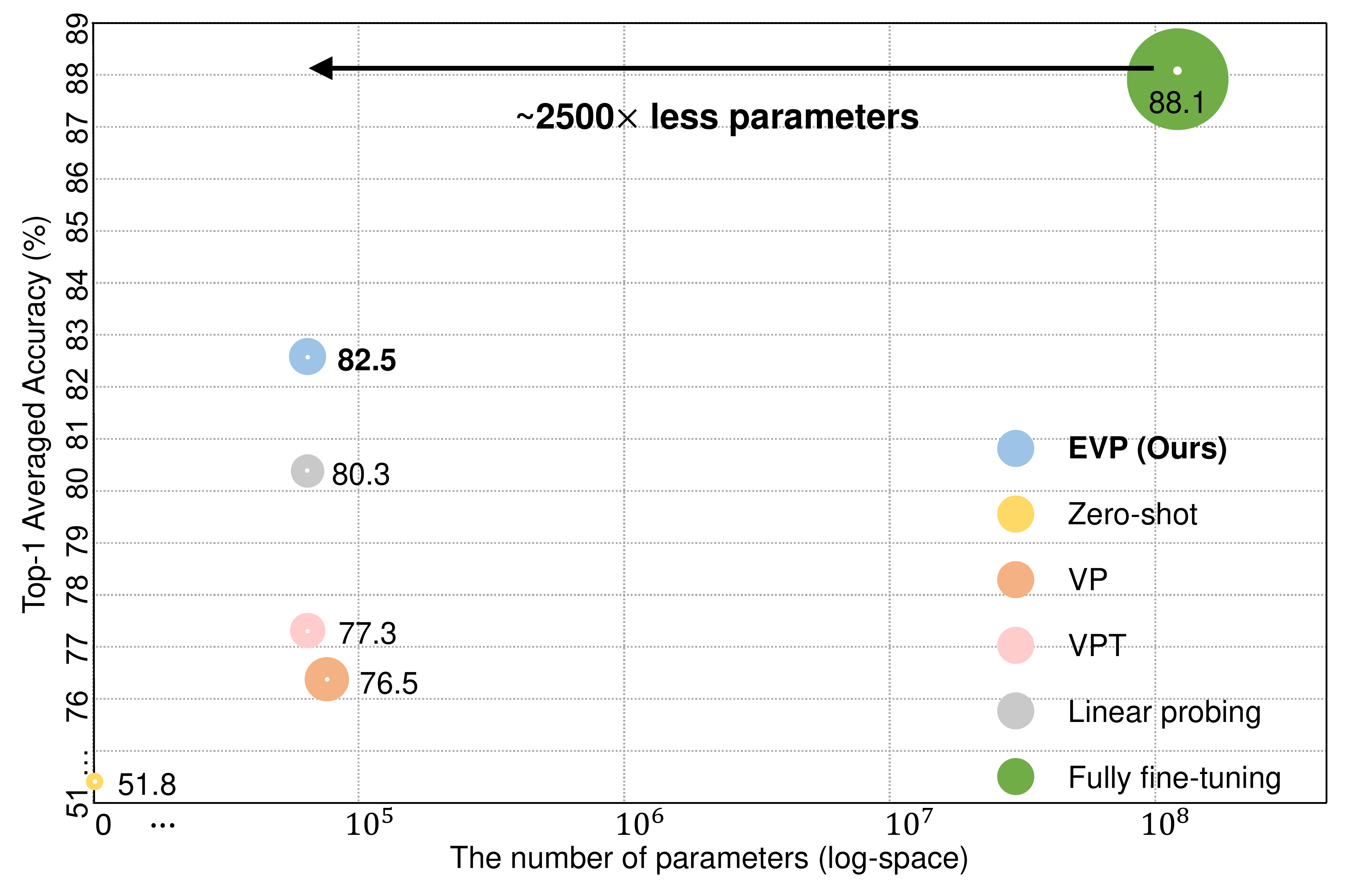}
\vspace{-1.5em}
\caption{The trade-off between the number of parameters and the average accuracy (across 12 datasets). Our method outperforms linear probing and other visual prompting baselines by a significant margin with a similar number of parameters.}
\vspace{-.5em}
\label{fig:intro_result} 
\end{figure}

\textit{Prompting} method, which only modifies the input space, offers an effective and efficient solution in NLP \cite{gao2021making, prompttuning, prefix}, \eg, text prompting can closely match the performance of fully fine-tuning \cite{ptune}. This promising result has motivated researchers to investigate whether similar success can be achieved in the field of computer vision. Some early efforts in this direction include VPT \cite{VPT} and VP \cite{VP}, which add small amounts of learnable parameters as tokens or perturbations directly at the pixel level to adapt foundation models. However, when taking a closer look at the trade-off between performance and parameter efficiency as shown in \cref{fig:intro_result}, we note these advanced visual prompting methods appear to be less competitive. For example, despite having a comparable number of parameters, there exists a significant performance gap between VPT and the simple linear probing baseline (\textbf{77.3\% \vs 80.3\%}). In this paper, \emph{we aim to unleash the full potential of visual prompting at the pixel level, and more importantly, to explore whether it can be stronger than other alternatives such as linear probing.}

Intriguingly, we find that with proper modifications, visual prompting can turn into a truly effective paradigm for adapting foundation models to different visual tasks. The first key observation is that directly adding together the prompt and the image, as in VP \cite{VP}, may corrupt the original image information, therefore limiting the learning of the prompt. We provide a simple strategy to resolve this issue: we first shrink the original image into a smaller size and then pad the prompt around it, \ie, the prompt and the image now do not overlap and are kept separate from one another. This strategy allows for independent optimization of the prompt, and enables flexibility in adjusting the padding size to control computation overheads. In addition, we draw on techniques from adversarial examples, which have similarities to visual prompting in their aim to either maximize or minimize the loss function \cite{VP,reprogram}, to further improve the performance of visual prompting. Specifically, we find that gradient normalization \cite{FGSM,dong2018boosting} and input diversity \cite{xie2019improving} are effective at improving the generalization ability of the prompt.

We follow the standard evaluation protocol to conduct experiments across 12 visual benchmarks. We note that, with a CLIP model \cite{CLIP}, our method attains an averaged accuracy of 82.5\%, significantly outperforming the previous state-of-the-art visual prompting method VPT \cite{VPT} by $\mathbf{+5.2\%}$. More excitingly, this 82.5\% result is +2.2\% stronger than the linear probing result (80.3\%) and even comparable to fully fine-tuning on certain datasets. We further confirm the superiority of our method in learning with data at different scales and in handling out-of-distribution samples. We hope our study can inspire further research in the field of visual prompt learning.
\section{Related Works}
\paragraph{Prompt learning in NLP.}
The key idea of prompting is to reformulate the input text in downstream tasks so that the frozen language models can better \textit{``understand''} and perform the task \cite{promptsurvey}. Prior works \cite{gpt3,petroni2019language, cui2021template} show that manually designed text prompt can help language models achieve remarkable representation capacity in the few-shot or even zero-shot settings at downstream tasks, but this requires specific domain knowledge. To address this issue, recent works have started to focus on prompt tuning \cite{prefix, ptune, prompttuning}, which involves directly optimizing the continuous prompting vector through gradient information. In this work, we investigate prompt learning in computer vision, which is a more challenging task because it involves a different type of signal (visual rather than language) that contains much less high-level semantic information.

\paragraph{Visual prompt learning.}
After witnessing the success of prompting in language models, researchers begin to explore the usage of prompts in the field of computer vision. For example, CoOp \cite{zhou2022learning} applies prompt tuning to vision-language models, learning the soft prompts through minimizing the classification loss on downstream tasks. VP \cite{VP} and VPT \cite{VPT} focus on prompting with images: VP optimizes the prompt directly in the pixel space, and VPT proposes to insert a set of learnable tokens into ViT architectures \cite{ViT} for prompt tuning.  While these approaches show the potential of visual-only prompt learning, as shown in \cref{fig:intro_result}, their performance is not as competitive compared to other methods such as linear probing. In this work, we aim to enhance visual prompt learning and demonstrate its strong potential for improving the performance of foundation models on a range of visual tasks.

\paragraph{Adversarial examples.}
It is well-known that machine learning models are vulnerable to adversarial attacks  \cite{advclassification, biggio2013evasion, huang2011adversarial}. The fast gradient sign method (FGSM) \cite{FGSM} and projected gradient descent (PGD) \cite{madry2018towards} are two commonly used techniques for creating adversarial examples that can fool deep learning models. Nonetheless, these adversarial examples cannot transfer well to fool other models. Later works show that the difficulty of optimizing adversarial examples is the cause of this weak transferability, and techniques such as diverse input patterns \cite{xie2019improving} and momentum-based gradient accumulation \cite{momentum_adv2017} have been proposed to improve transferability. Given the similarity between generating adversarial examples and the process of prompt learning \cite{VP,reprogram}, we hereby are interested in revisiting techniques from building transferable adversarial examples to enhance visual prompting.

\section{Methodology}
In this section, we present \textbf{E}nhanced \textbf{V}isual \textbf{P}rompting (EVP), a simple and effective pixel-level visual prompting method for adapting foundation models to downstream tasks. We first provide a thorough review of previous visual prompting methods as preliminaries, including VP and VPT, and then describe the prompting design and training strategy of our EVP in detail.

\begin{figure*}[t!]
\centering
\includegraphics[width=0.88\linewidth]{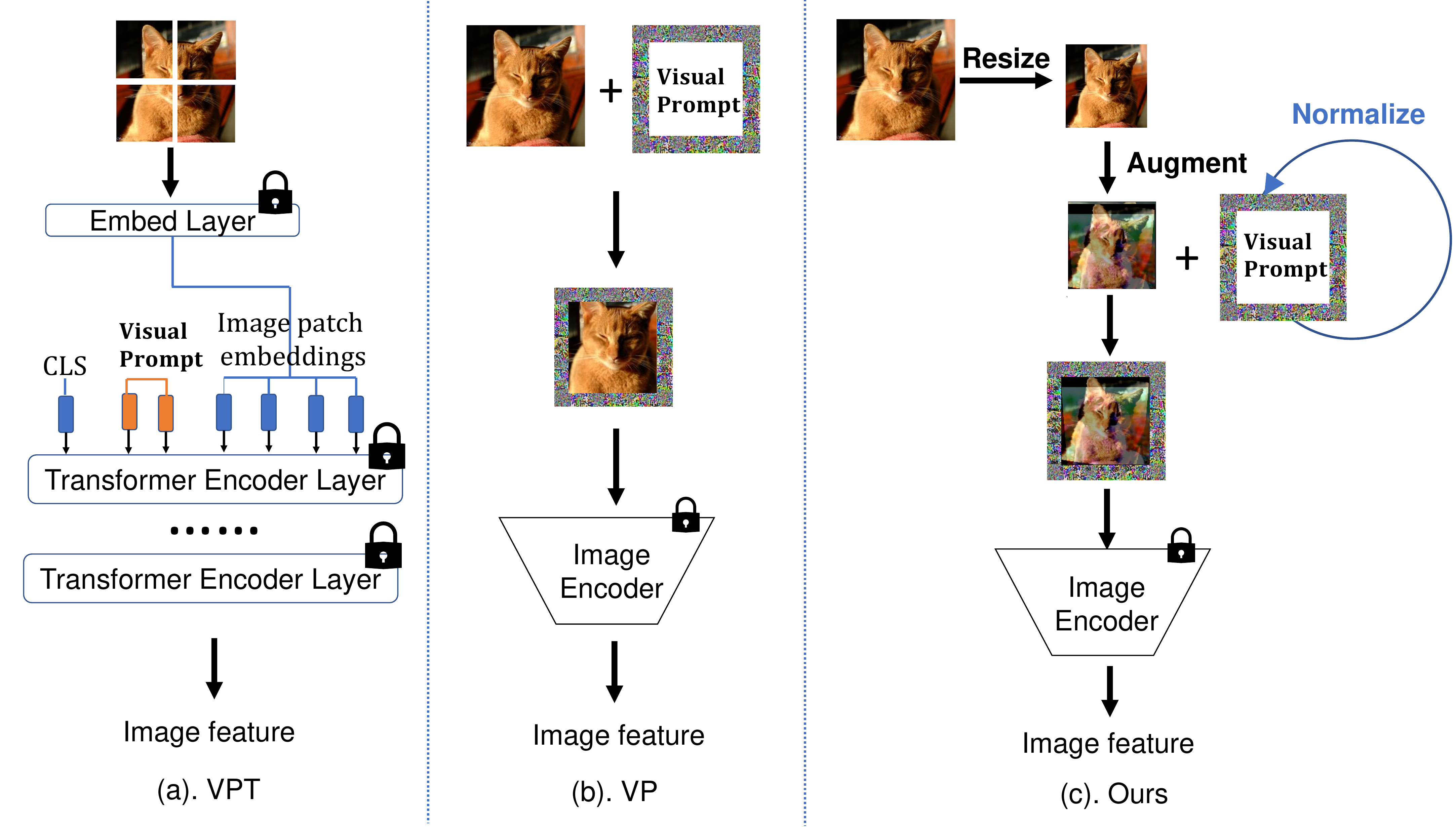}
\caption{\textbf{Overview of the designs of different visual prompting methods.} (a) VPT: Injecting a set of learnable parameters into the token space; (b) VP: Modifying learnable perturbations on the border of input images; (c) Our EVP: Shrinking images and applying data augmentations, then padding the prompt around the image. Note the prompts in EVP are updated using normalization strategies inspired by adversarial attack techniques \cite{FGSM}.}
\label{fig:method}
\vspace{-.5em}
\end{figure*}

\subsection{Preliminaries}
\paragraph{VPT \cite{VPT}} adds a set of learnable parameters into ViT architecture for visual prompting. For a fair comparison, we hereby consider its VPT-SHALLOW version, which only inserts the prompt into the first layer’s input. Specifically, as shown in \cref{fig:method} (a), VPT inserts a collection of prompts $P$ between the learnable class token \texttt{[CLS]} and a sequence of patch embeddings $E$, creating a new input $x = [\texttt{CLS}, P, E]$.

\paragraph{VP} \cite{VP} adapts foundation models to downstream tasks by directly adding together the learnable prompt and the input images at the pixel level. The prompt $v_{\phi}$ is designed to be input-agnostic and task-specific, and is placed on the border of the input images, as shown in \cref{fig:method} (b). 
During training, VP maximizes the likelihood of the correct label $y$ by optimizing the prompt $v_\phi$: $\max_{v_\phi} P (y | x + v_{\phi})$. During inference, the optimized prompt is then added to test images: $X_{test} = \{x^{1}_{test} + v_{\phi}, \ldots,  x^{n}_{test} + v_{\phi} \}$.

\subsection{Designing EVP}
Our prompting design is largely based on VP, but with some simple modifications. The issue we identify with VP is that directly adding together the prompt and the images may corrupt the original image information. For example, in \cref{fig:method} (b), the cat ears are heavily overlapped and obscured by the added prompts. This could hinder the learning of prompts (see our ablations in \cref{sec:image_size}). To address this issue, as shown in \cref{fig:method} (c), our EVP shrinks the input images and pads the prompt around them. Specifically, for an input image $X$\,$\in$\,$\mathbb{R}^{K\times K\times 3}$, it is shrunk to $\hat{x}$\,$\in$\,$\mathbb{R}^{k\times k\times 3}$ and then padded with $(K^{2}$\,$-$\,$k^{2})$\,$\times$\,$3$ prompts to obtain the output image $\hat{X}$\,$\in$\,$\mathbb{R}^{K\times K\times 3}$. Similar to VP, during training we optimize the prompt by maximizing the likelihood of the correct label, and during inference we pad the optimized prompt around the shrunk test samples for predictions.

It is important to note that while both EVP and VPT keep the prompt and the image non-overlapping, there is a key difference between these two methods. Specifically, the prompts in EVP are later added with positional embedding, while this is not the case for VPT. As shown in the ablation study in \cref{positon_ablation}, positional information is crucial for achieving strong performance with visual prompting.

\subsection{Training Strategy of EVP}
There is a strong relationship between prompting and adversarial attacks. In adversarial attacks, the goal is to learn a pixel perturbation $g_i$ that will mislead the network given an image $x_i$. This can be formulated as $\min_{g_{i}} P(y_{i} | x_{i} + g_{i})$. On the other hand, visual prompting can be seen as the inverse process of adversarial attacks, in which the aim is to learn a template $v$ that will maximize the likelihood of the correct label $y$. Given this relationship, we are motivated to explore whether techniques from adversarial attacks, particularly those focused on building transferable adversarial examples, can be useful for visual prompting.

\paragraph{Input diversity.}
Previous work \cite{xie2019improving} demonstrates that input diversity can help optimization and improve the transferability of adversarial examples. As shown in later works, the concept of ``diverse input'' can be generalized to apply different data augmentation strategies in generating adversarial examples \cite{dong2019evading,wang2021admix,wu2021improving}. We hereby re-introduce this concept into visual prompting. Specifically, our EVP considers a range of augmentation including RandomHorizontalFlip, RandAug~\cite{cubuk2020randaugment}, and Cutmix~\cite{yun2019cutmix}. As shown in \cref{sec:training}, we find that simple augmentation strategies like \texttt{RandomHorizontalFlip} are already sufficient to significantly improve visual prompting.

\begin{table*}[t!]
\caption{Performance comparison across 12 datasets with CLIP. We note EVP substantially outperforms other visual prompting methods by a large margin. More notably, EVP outperforms the linear probing on 7 out of 12 datasets with a similar number of parameters. The results where EVP outperforms linear probing are highlight in \textbf{bold}.}
\vspace{.2em}
\label{table:CLIP reuslts}
\resizebox{\linewidth}{!}{
\begin{tabular}{ccccccccccccccc}
\toprule
 \multicolumn{1}{c|}{Adaptation} & \multicolumn{1}{c|}{Tunable params (M)}& \multicolumn{1}{c|}{CIFAR100} & \multicolumn{1}{c|}{CIFAR10} & \multicolumn{1}{c|}{Flowers} & \multicolumn{1}{c|}{Food} & \multicolumn{1}{c|}{EuroSAT} & \multicolumn{1}{c|}{SUN}&\multicolumn{1}{c|}{DMLab} & \multicolumn{1}{c|}{SVHN} & \multicolumn{1}{c|}{Pets} & \multicolumn{1}{c|}{DTD} & \multicolumn{1}{c|}{RESISC}&\multicolumn{1}{c|}{CLEVR}&\multicolumn{1}{c}{Avg.} \\ 
\midrule
\midrule
 TP          &0                    & 63.1                          & 89.0                         & 61.8                         &83.2                      & 34.1                         & 58.0   &30.2                  & 11.0                       & 85.9                      & 42.8                     & 42.4             &20.2                & 51.8                         \\ \hline
 VP         &0.070                  & 75.3                          & 94.2                         & 62.0                         & 83.2                      & 95.6                         & 68.4   &41.9                    & 88.4                      & 86.5                      & 57.1                     & 84.1            &81.4                   & 76.5                         \\ \hline
 VPT       &0.064                    & 76.6                          & 95.0                         & 76.2                         & 84.7                      & 94.6                         & 69.3      & 48.4               & 86.1                     & 92.1                      & 61.6                     & 84.3                    &58.6                    &77.3              \\ \hline

 
\rowcolor{mygray}
 EVP(Ours)       &0.062               & \textbf{81.2}                         & \textbf{96.6}                & 82.3                         & 84.1             & \textbf{97.6}                & 71.0   &\textbf{62.3}                  & \textbf{90.5}                      & \textbf{90.0}             & 68.4                     & 89.7         &\textbf{75.9}                    & 82.5                         \\ \hline
 LP        &0.037                      & 80.0                          & 95.0                         & 94.1                         & 88.3                      & 94.8                         & 76.2   &49.3         & 65.4                      & 89.2                      & 73.5           & 92.3           &66.1             & 80.3                         \\ \hline
 FT        &151.28                      & 82.1                 & 95.8                & 97.4                & 87.8                      &99.0                            & 79.0           &63.5             & 95.7            & 88.5                      & 72.3                     &98.1      & 94.4                  & 88.1                         \\ 
\bottomrule
\end{tabular}
}
\vspace{-.2em}
\end{table*}

\begin{figure*}[t!]
\centering
\includegraphics[width=.9\linewidth]{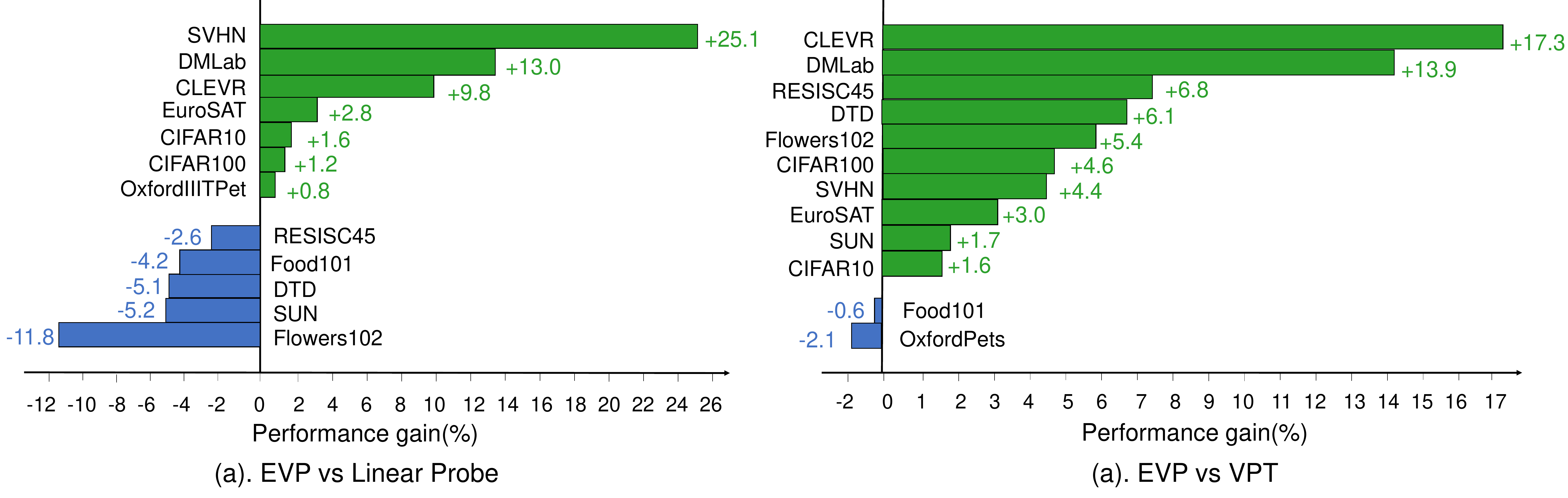}
\caption{\textbf{Performance gain} of EVP compared to linear probing and VPT on each downstream dataset. The bars indicate the gain or loss in accuracy compared to linear probing and VPT, respectively. (a) Compared with linear probing, EVP outperforms linear probing on 7 out of 12 datasets by \textbf{2.1\%} on average. (b) Compared with VPT, EVP beats  VPT on 10 out of 12 datasets by \textbf{5.6\%} on average. }
\label{fig:main_result} 
\end{figure*}

\paragraph{Gradient normalization.}
In adversarial attacks, it is common to apply normalization techniques, such as the $L_1$, $L_2$, or $L_\infty$ norm, to the gradients update \cite{FGSM}. For example, using the $L_2$ norm, the gradient can be normalized as follows:
\begin{equation}
    x^{adv} = x + \gamma \frac{\nabla_{x} J(x,y)}{||\nabla_{x} J(x,y)||_{2}} \,,   \label{adv}
\end{equation}
where $\gamma$ is the learning rate, $J$ is the loss function, and $\nabla_{x} J$ is the gradient of the loss function \wrt the input $x$.

We hereby introduce this gradient normalization to visual prompting. We define the matrix representation of EVP as $V_{e} = W \odot M$, where $W\in \mathbb{R}^{K\times K\times 3}$ are the prompt parameters, $M\in \mathbb{R}^{K\times K\times 3}$ is the mask matrix, and $\odot$ denotes the element-wise matrix product. The mask matrix $M$ encodes the spatial locations of the prompts, with the central part of size $k \times k$ being all zeros and the rest being all ones. In practice, we find that dividing the gradient of EVP by the $L_{2}$ norm of the gradient of $W$ leads to the best performance:
\begin{equation}
    V_{e}^{t+1} = V_{e}^{t} - \gamma \frac{\nabla 
_{V_{e}^{t}} J}{||\nabla_{W} J||_{2}}\, .
\end{equation}
We provide more details and ablation results on different normalization strategies in \cref{sec:training}.
\section{Experiments}
\label{sec:experiments}

\paragraph{Datasets.} We evaluate visual prompting methods on 12 downstream classification datasets, including CIFAR100, CIFAR10 \cite{krizhevsky2009learning}, Flowers102 \cite{nilsback2008automated}, Food101 \cite{bossard2014food}, EuroSAT \cite{helber2019eurosat}, SUN397 \cite{xiao2010sun}, SVHN \cite{netzer2011reading}, DTD \cite{cimpoi2014describing}, OxfordPets \cite{parkhi2012cats}, Resisc45 \cite{cheng2017remote}, CLEVR \cite{johnson2017clevr}, and DMLab \cite{beattie2016deepmind}. In addition, we test the robustness of visual prompting on 3 out-of-distribution datasets \cite{koh2021wilds} (Camelyon17, FMoW, and iWildCAM), and 2 corruption datasets \cite{hendrycks2018benchmarking} (CIFAR100-C and CIFAR10-C). 

\paragraph{Baselines.} We compare the performance of EVP with other commonly used prompting methods and fine-tuning protocols, including TP (text prompting), VP, VPT, LP (linear probing), and FT (fully fine-tuning). Specifically, we should note 1) TP is equivalent to zero-shot in CLIP; 2) LP uses a linear layer as the classification head; and 3) FT updates all parameters of the backbone and the classification head.

\begin{table*}[t!]
\caption{\textbf{Performance of non-CLIP models}. EVP\textsuperscript{*} indicates that we train EVP using classes after preprocessing stage. EVP slightly exceeds VP and EVP\textsuperscript{*} outperforms EVP and VP by a large margin. The \textbf{bold} indicates the cases that the performance of EVP\textsuperscript{*} is competitive with linear probing.}
\vspace{.2em}
\label{table:other models}
\resizebox{\linewidth}{!}{
\begin{tabular}{cccccccccccccc}
\toprule
\multicolumn{1}{c|}{Model} & \multicolumn{1}{c|}{Adaptation} & \multicolumn{1}{c|}{CIFAR100} & \multicolumn{1}{c|}{CIFAR10} & \multicolumn{1}{c|}{Flowers} & \multicolumn{1}{c|}{Food} & \multicolumn{1}{c|}{EuroSAT} & \multicolumn{1}{c|}{SUN} & \multicolumn{1}{c|}{SVHN} & \multicolumn{1}{c|}{Pets} & \multicolumn{1}{c|}{DTD} & \multicolumn{1}{c|}{RESISC}&\multicolumn{1}{c|}{CLEVR}&\multicolumn{1}{c}{Avg.} \\ 
\midrule
\midrule

Instagram                        & VP                           & 16.7                          & 62.1                         & 4.8                         & 6.5                      & 86.1                        & 2.2                     & 53.8                      & 18.6                      & 29.1                     & 40.6            &30.9         & 31.9                          \\ 

Instagram                        & EVP                           &      13.6                     &  67.2                       &     9.2                     &      7.1                 & 85.6                        & 7.9                     &    50.8                   & 16.3                      &  29.0                    &    38.0        &   48.1      & 33.9                         \\ 

\rowcolor{mygray}
Instagram                        & EVP\textsuperscript{*}                     &     \textbf{60.3}                     &     \textbf{93.5}          &11.4                          & 8.4             &      88.7          &19.6                          & \textbf{ 55.3}                       & 74.4           & 44.4                     &47.5        & \textbf{50.5}           &50.4                         \\ 
Instagram                        & LP                              & 64.0                          & 90.1                         & 92.7                         & 65.8                     & 95.5                        & 58.1            & 48.0                      & 94.5                      & 70.9            & 95.7           &30.2           & 73.2                        \\ 
Instagram                        & FT                              & 77.8                 & 77.8               & 94.5                &75.6                      & 97.4                & 56.7                    & 96.8             & 93.9                      &73.5                     & 93.4     &87.9       & 84.1                         \\ \hline

RN50                        & VP                           & 10.1                          & 54.5                         &4.7                          & 5.1                      & 80.7                        & 1.1                     & 57.1                      & 10.8                      & 8.2                     & 28.3            &29.5         & 26.4                         \\ 

RN50                        & EVP                           &  9.2                         &   55.9                       &     6.6                     & 3.9                      &  85.5                       &   5.1                   &     48.6                  & 10.5                       &  18.7                    &  35.4           &  35.5       &     28.6                     \\ 

\rowcolor{mygray}
RN50                        & EVP\textsuperscript{*}                     &24.9                          & 77.0              &  11.9                        &7.0              & 81.0                & 14.7                     & 47.8                      &72.0            &41.2                      &39.2        &\textbf{37.2}           & 41.3                        \\ 
RN50                        & LP                              & 67.7                          & 87.7                         & 92.7                         & 62.5                    & 95.8                         & 57.5            & 60.3                      & 91.1                      & 66.7            & 92.2           &32.6           & 73.3                        \\ 
RN50                        & FT                              &79.9                 & 94.1              & 96.9                &73.2                      & 96.5                & 55.9                    & 96.9             & 92.3                      &66.7                     &93.4     &89.3       & 84.3                         \\

\bottomrule
\end{tabular}
}
\end{table*}

\begin{table*}[t]
\caption{ Robustness comparison on \textbf{out-of-distribution} and \textbf{corruption} datasets. Left: out-of-distribution datasets. Right: corruption datasets. We can observe that EVP achieves much stronger robustness on both out-of-distribution setting and corruption setting.}
\vspace{0.2em}
\begin{minipage}{\textwidth}
\begin{minipage}[b]{0.48\textwidth}

\label{table: OOD test}
\resizebox{\linewidth}{!}{
\begin{tabular}{cccccc}
\toprule
\multicolumn{1}{c|}{Model} & \multicolumn{1}{c|}{Adaptation} &  \multicolumn{1}{c|}{iwildcam} & \multicolumn{1}{c|}{camelyon17} & \multicolumn{1}{c|}{fmow}&\multicolumn{1}{c}{Avg.} \\ 
\midrule
\midrule
CLIP                        & VP                                                                               &57.3                        & 91.4                     & 37.8                         &   62.2                     \\ \hline
CLIP                        & VPT                                                                               &58.8                       & 91.9                   &29.7                        &60.1                   \\ \hline
\rowcolor{mygray}
CLIP                        & Ours                                                         &64.9                         &\textbf{95.1}              &40.2                    &  \textbf{66.7}                       \\ \hline
CLIP                        & LP                                                                            &\textbf{66.7}                        &86.0                      &36.3                         &63.0       \\ \hline
CLIP                        & FT                              & 64.0                &84.3                       &\textbf{49.7}                & 66.0                      \\ 
\bottomrule
\end{tabular}
}

\end{minipage}
 \hfill
\begin{minipage}[b]{0.48\textwidth}

\label{table: corruption test}
\resizebox{0.94\linewidth}{!}{
\begin{tabular}{ccccc}
\toprule
\multicolumn{1}{c|}{Model} & \multicolumn{1}{c|}{Adaptation} & \multicolumn{1}{c|}{CIFAR100-C} & \multicolumn{1}{c|}{CIFAR10-C} &\multicolumn{1}{c}{Avg.} \\ 
\midrule
\midrule
CLIP                        & VP                           &     52.5
                      &       78.3                          &     65.4                   \\ \hline
CLIP                        & VPT                           &   54.0                        & 70.2                                              &    62.3               \\ \hline
\rowcolor{mygray}
CLIP                        & Ours                      & 58.6                        &\textbf{84.3}                                  &    71.5                     \\ \hline
CLIP                        & LP                              &56.9                          &78.8                                                &   67.9    \\ \hline
CLIP                        & FT                              &   \textbf{61.1}              & 82.7                        &            71.9           \\ 
\bottomrule
\end{tabular}
}
\end{minipage}
\end{minipage}
\end{table*}

\subsection{CLIP}
\label{subsec: clip}
Following the protocol of VP \cite{VP}, we conducted evaluations using the CLIP-Base/32 model on 12 classification datasets. The full results are shown in \cref{table:CLIP reuslts} and a detailed comparison to the two strong baselines, LP and VPT, is presented in \cref{fig:main_result}. Our proposed EVP approach consistently outperformed all previous prompting methods, with similar or fewer parameters. On average, EVP showed an improvement of 6.0\% over VP and 5.4\% over VPT.

To further evaluate the effectiveness, we next compare EVP with linear probing, which is a widely used fine-tuning protocol. The results, shown in \cref{table:CLIP reuslts} and \cref{fig:main_result}, demonstrate that EVP outperforms linear probing on 7 out of 12 datasets. On average, EVP achieved an accuracy of 82.5\%, which is 2.2\% higher than linear probing. In addition, our method is more flexible as the number of parameters can be easily controlled (via adjusting the padding size), whereas the number of parameters in linear probing must depend on the number of class categories in the downstream tasks.
    
Lastly, Our method, EVP, exhibits promising performance compared to fully fine-tuning while being significantly more parameter-efficient, with only 0.04\% of the number of parameters. While there is still a performance gap between these two methods, with an average accuracy of 82.5\% for EVP and 88.1\% for fully fine-tuning, EVP outperforms or achieves similar results on certain datasets, including CIFAR100, CIFAR10, EuroSAT, DMLab, and Pets.

\subsection{Non-CLIP Models}
In this section, we evaluate the effectiveness of EVP on non-CLIP models. One challenge for adapting non-CLIP models to downstream tasks is that their original classification head is either less semantically meaningful or mapped to a set of predefined classes. A direct and naive solution used in VP is to arbitrarily map downstream classes to pre-trained classes and discard any unassigned classes. However, we posit that there could exist some similarity between pre-trained and downstream classes, even if an exact correspondence is not known. This motivates us to propose a pre-processing stage before implementing visual prompting to utilize this potential similarity.

Specifically, for each downstream class, we feed downstream images in that class into the pre-trained model and investigate the predictions in the pre-trained classes. We then simply choose the pre-trained class with the highest prediction frequency as the corresponding class for the downstream class. After pre-processing, we fix the correspondence and train our visual prompting method.

Overall, the results in \cref{table:other models} demonstrate the effectiveness of our proposed EVP method for adapting non-CLIP models to downstream tasks. While the performance of EVP is already improved over the baseline VP method (by $\sim$ 2\%) when using arbitrary mapping, the use of our pre-processing stage substantially enhances the performance of EVP further, \ie, from 33.9\% to 50.4\% with the Instagram pretraining model \cite{mahajan2018exploring} and from 28.6\% to 41.3\% with an ImageNet-train ResNet-50. However, we note the performance of EVP\textsuperscript{*} is not as strong on fine-grained datasets, such as Flowers102 and Food101, suggesting that it may be more challenging to find correspondence between pre-trained and downstream classes for these types of tasks.

\begin{figure*}[t!]
\centering
\includegraphics[width=.98\linewidth]{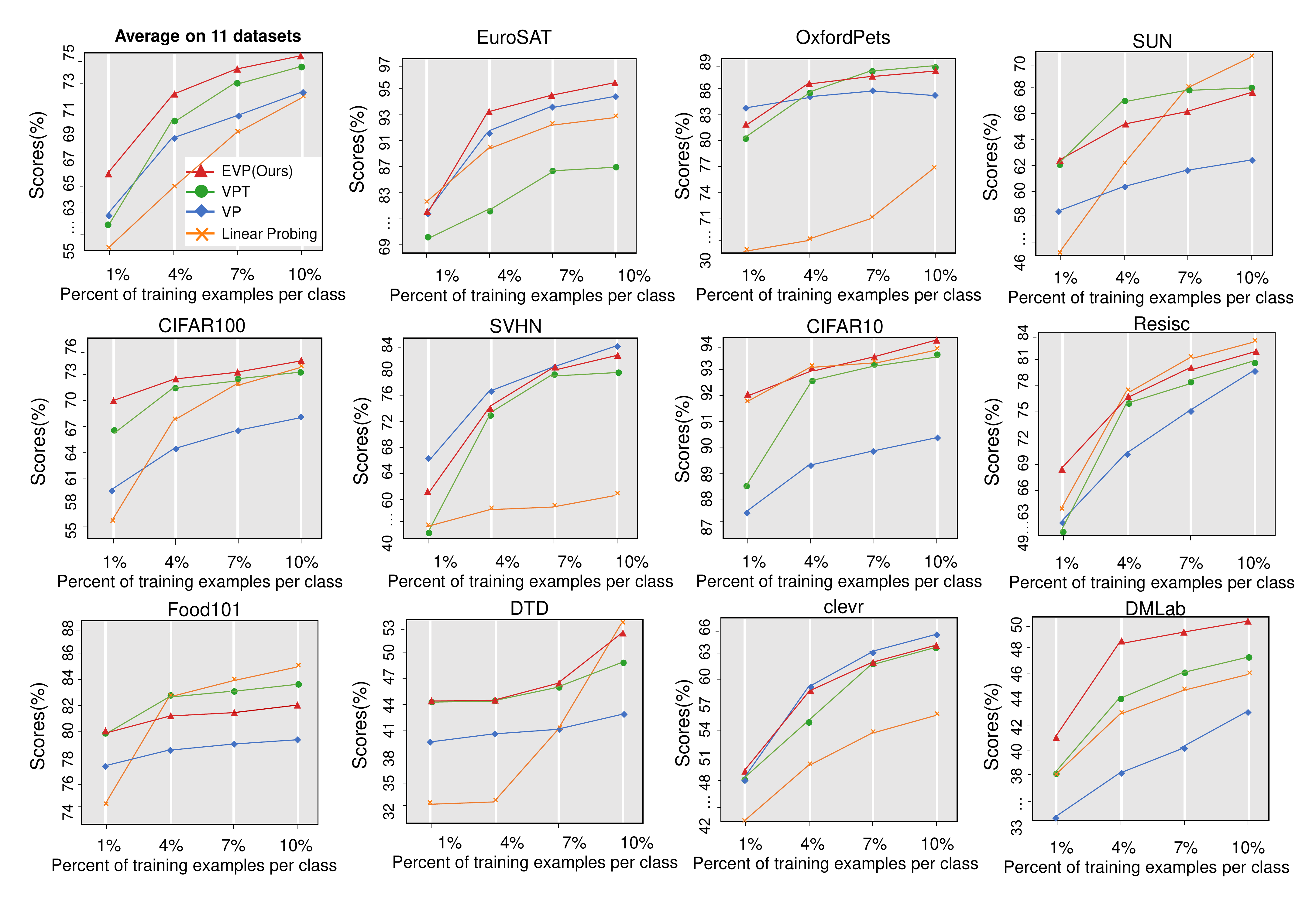}
\vspace{-.5em}
\caption{\textbf{Results with different data scales} on 11 visual recognition datasets. Each figure shows the results trained on 1\%, 4\%, 7\%, 10\% data, respectively. All visual prompting methods show clear dominance compared with linear probing. EVP (red line) outperforms other methods by a large margin on average.}
\vspace{-0.2em}
\label{fig:fewshot} 
\end{figure*}

\begin{table*}[t!]
\caption{\textbf{Ablation on original image information} across 12 datasets on CLIP. There is a clear trend that as the image size decreases, \ie, as the occlusion decreases, the performance gradually increases.}
\vspace{0.2em}
\label{table:origin_info}
\resizebox{\linewidth}{!}{
\begin{tabular}{cccccccccccccc}
\toprule
 \multicolumn{1}{c|}{Image size} & \multicolumn{1}{c|}{CIFAR100} & \multicolumn{1}{c|}{CIFAR10} & \multicolumn{1}{c|}{Flowers} & \multicolumn{1}{c|}{Food} & \multicolumn{1}{c|}{EuroSAT} & \multicolumn{1}{c|}{SUN}&\multicolumn{1}{c|}{DMLab} & \multicolumn{1}{c|}{SVHN} & \multicolumn{1}{c|}{Pets} & \multicolumn{1}{c|}{DTD} & \multicolumn{1}{c|}{RESISC}&\multicolumn{1}{c|}{CLEVR}&\multicolumn{1}{c}{Avg.} \\ 
\midrule
\midrule
 224                              & 77.1                          & 94.7                         & 80.6                        &81.8                      & 96.7                &71.4   &60.1                  & 88.9                       & 89.1                      & 64.2                     & 88.1             &73.4                & 80.5                         \\ \hline
 204                           & 77.5                          & 95.3                         & 80.8                         & 82.0                      & 97.0                         & 71.0   &60.7                    & 89.7                      & 89.3                      & 64.0                     & 88.5            &73.6                   & 80.8                         \\ \hline
 184                           & 79.0                          & 95.9                         & 81.7                         & 82.3                      & 97.4                         & 70.3      & 61.9               & \textbf{90.6}                     & \textbf{89.9}                      & 65.2                     & 89.5                    &74.3                    &81.5              \\ \hline
\rowcolor{mygray}
 164 (Default)                      & \textbf{81.2}                         & \textbf{96.6}                & \textbf{82.3}                         & \textbf{82.3}             & \textbf{97.6}                & 71.0   &\textbf{62.3}                  & 90.5                      & 88.7             & \textbf{68.4}                     & \textbf{89.7}         &\textbf{75.9}                    & 82.2                         \\ 
\bottomrule
\end{tabular}
}
\vspace{-1em}
\end{table*}

\subsection{Robustness}

In this section, we investigate the robustness of EVP compared with other prompting methods and fine-tuning protocols. We evaluate the robustness of EVP on out-of-distribution (OOD) and corruption datasets.

\paragraph{OOD robustness.} We test the robustness of EVP to distribution shift using the WILDS benchmark \cite{koh2021wilds}. The model is trained on datasets from a specific domain and then evaluated on datasets from a different domain, such as images from different regions, cameras, and hospitals. The results in \cref{table: OOD test} show that EVP outperforms other prompting methods by at least 4.5\%. Additionally, we find that EVP outperforms both linear probing (+3.7\%) and fully fine-tuning (+0.7\%) in this setting, highlighting the potential of EVP in handling out-of-distribution samples.

\paragraph{Robutness on corruption datasets.}
In this study, we also evaluate the robustness of EVP to common image corruptions \cite{hendrycks2018benchmarking}. We test EVP on the CIFAR100-C and CIFAR10-C corruption datasets, which apply 19 common image corruptions to the CIFAR100 and CIFAR10 datasets, respectively. We train EVP on the CIFAR100 and CIFAR10 datasets and then evaluate its performance on the corresponding corruption datasets. 

The average accuracy is reported in \cref{table: corruption test} (the accuracy under each type of corruption is reported in the supplementary material), where we can observe that EVP outperforms other prompting methods and linear probing by a large margin. It is also worth noting that EVP performs comparably to fully fine-tuning in handling common image corruptions, \ie, 71.5\% \vs 71.9\%. This may be due to the fact that the corruptions on the image can damage the performance of other baselines, while our strategy of treating the prompt as a standalone and independent component can alleviate this issue, \ie, the prompts in EVP are not directly combined with (but are padded around) the corrupted images during inference.

\begin{figure*}[t]
\centering
\includegraphics[width=.93\linewidth]{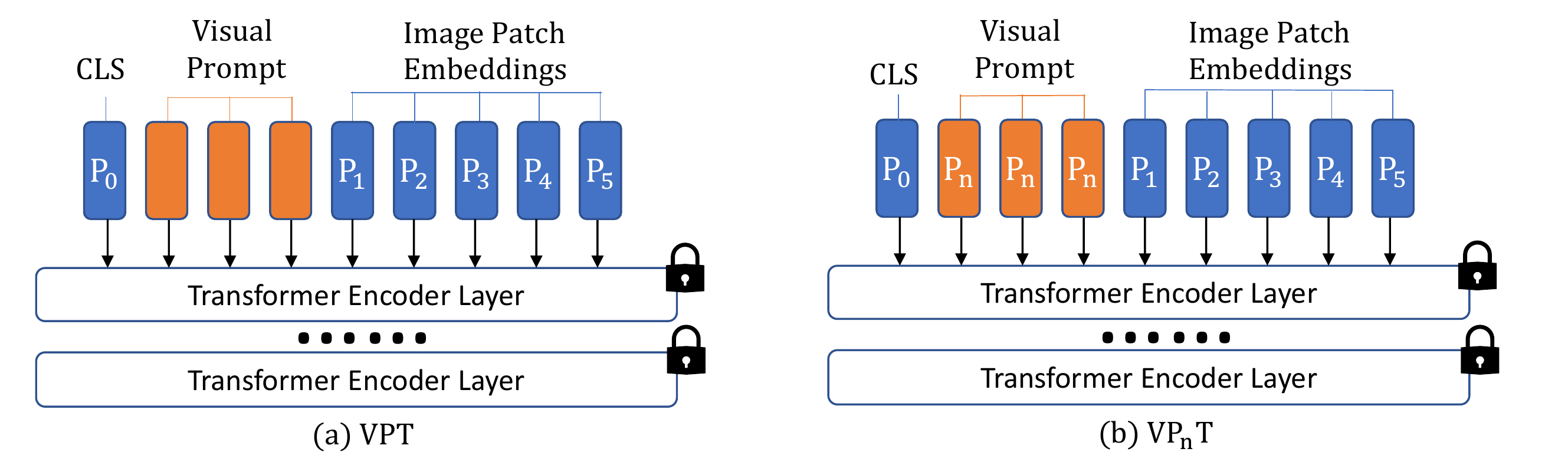}
\vspace{-.2em}
\caption{\textbf{Ablation on positional embedding at token level.} (a). Visual Prompting Tuning(VPT): Inject learnable tokens between CLS and image patch embedding without positional embedding (b): V$\text{P}_{n}$T: Inject learnable tokens between CLS and image patch embedding with same positional embedding $\text{P}_{n}$ (i.e, \textit{n}-th positional embedding (n = 1, 2, \ldots, 5)).
}
\vspace{-.3em}
\label{fig:VPT-head}
\end{figure*}

\subsection{Different Data Scales}
\label{sub:data scale}
In this section, we evaluate the performance of EVP with different data scales. We train EVP using only 1\%, 4\%, 7\%, and 10\% of the data for each class in the training datasets. This is of particular interest because few-shot learning is an important aspect of prompting in natural language processing. We hereby aim to validate if EVP can achieve strong performance with limited data.

The average accuracy, as well as the detailed accuracy on each dataset, is reported in \cref{fig:fewshot}. We can observe that 1) visual prompting methods (VP, VPT, EVP) consistently outperform linear probing, demonstrating the effectiveness of visual prompting in learning with limited labeled data; and 2) among all visual prompting methods, EVP consistently achieves the best overall performance, demonstrating its strong generalization ability at different data scale. 

\section{Ablation Study}
\label{sec:ablation}

\subsection{Original Image Information}
\label{sec:image_size}
We first investigate the importance of preserving the original information of the input image. To do this, we vary the size of the input image while keeping the number of learnable parameters constant, \ie, corrupt the border information of the original image to different degrees. The results are presented in \cref{table:origin_info}. We can see that the performance of 9 out of 12 datasets increases as the degree of corruption decreases, indicating the importance of preserving original image information.

\subsection{Prompting Positional Embedding}
\label{positon_ablation}
We next investigate the impact of positional embeddings. Note while EVP encodes the positions of both learnable visual prompts and image patch embeddings, VPT only includes positional embeddings (\textbf{PE}) for the image patch embeddings. We will compare different configurations to demonstrate the importance of positional embeddings for visual prompting, at both the pixel and token levels.

At the pixel level, we denote our main method as \textbf{EVP-small w/ PE}, and define two additional variations: 1) \textbf{EVP-big w/ PE}: We pad the original image with learnable pixels and interpolate the original positional embeddings to the appropriate size, then we add the positional embeddings to both image patch embeddings and learnable pixels. 2) \textbf{EVP-big w/o PE}: We pad the original image with learnable pixels and only add positional embeddings to the image patch embeddings. \cref{table: pixel PE} shows the results. 
We can observe \textbf{EVP-big w/ PE} achieves the best performance, while \textbf{EVP-big w/o PE} performs the worst. For example, we note even \textbf{EVP-small w/ PE} is able to outperform \textbf{EVP-big w/o PE} by an average of 6.1\% on the five datasets (86.7\% \vs 78.6\%), despite having fewer parameters and lower resolution. These results demonstrate the vital role of adding positional embeddings in visual prompting.


\begin{table}[t!]
\caption{\textbf{Ablation on the positional embedding at the pixel level}. EVP-small shrinks the original image and pads it with learnable pixels to the original size, while EVP-big pads pixel patches around the original image. We note EVP-small w/ PE can beat EVP-big w/o PE despite having fewer parameters and a smaller input resolution, suggesting positional embeddings are crucial in visual prompting.}
\vspace{0.2em}
\label{table: pixel PE}
\resizebox{1\linewidth}{!}{
\begin{tabular}{ccccccc}
\toprule
\multicolumn{1}{c|}{Methods} & \multicolumn{1}{c|}{CIFAR100} &  \multicolumn{1}{c|}{CIFAR10} & \multicolumn{1}{c|}{DTD} & \multicolumn{1}{c|}{RESISC}&\multicolumn{1}{c|}{EuroSAT}&\multicolumn{1}{c}{Avg.} \\ 
\midrule
\midrule
EVP-small w/ PE                     &81.2                   & 96.6                & 68.4                   & 89.7                          &97.6          & 86.7           \\
\midrule
EVP-big w/ PE                      & 81.4                  &96.9                & 68.9                     & 91.6                          &97.4           & 87.2      \\
\midrule
EVP-big w/o PE                     &73.4                    &93.7               & 64.6                     &76.2                        &85.3            & 78.6     \\ 

\bottomrule

\end{tabular}
}
\vspace{-.5em}
\end{table}

At the token level, we find that simply adding positional embeddings to the learnable tokens can improve performance. To investigate this further, we next design different prompting choices at the token level by adding different positional embeddings to the learnable tokens. Specifically, we denote prompting choices as V$\text{P}_{n}$T, where the \textit{n}-th positional embeddings are added to the learnable tokens, as illustrated in \cref{fig:VPT-head}. 
Specifically, in our experiment, we choose $n=\{1, 25, 50\}$ to study the effects of inserting tokens at the head, the middle, and the tail of the original image (which will result in 50 tokens in total).

The results in \cref{table: token PE} demonstrate that the incorporation of positional embeddings in the learnable tokens can consistently and significantly improve the performance of visual prompting. For example, the average accuracy is increased by +2.4\% with V$\text{P}_{1}$T, +1.8\% with V$\text{P}_{25}$T, and + 1.4\% with V$\text{P}_{50}$T. These results altogether confirm the importance of positional embeddings in this context.

\begin{table}[t!]
\caption{\textbf{Ablation on the positional embedding at the token level}. VPT only adds positional embeddings to the image patch embeddings, while V$\text{P}_{1}$T, V$\text{P}_{25}$T, V$\text{P}_{50}$T denote methods in which the \nth{1}, \nth{25}, and \nth{50} positional embeddings are added to the learnable tokens, respectively. We can observe that simply adding positional embeddings to the learnable tokens can significantly improve performance.}
\vspace{0.2em}
\label{table: token PE}
\resizebox{1\linewidth}{!}{
\begin{tabular}{ccccccc}
\toprule
\multicolumn{1}{c|}{Methods} & \multicolumn{1}{c|}{CIFAR100} &  \multicolumn{1}{c|}{CIFAR10} & \multicolumn{1}{c|}{DTD} & \multicolumn{1}{c|}{RESISC}&\multicolumn{1}{c|}{EuroSAT}&\multicolumn{1}{c}{Avg.} \\ 
\midrule
\midrule
VPT        &76.6                   & 95.0                & 61.6                   & 84.3                     &  94.6          & 82.4           \\
\midrule
V$\text{P}_{1}$T  & \textbf{77.3}                  &\textbf{96.0}                 & \textbf{67.7}                     &\textbf{87.0}                  &  \textbf{96.2}           & \textbf{84.8}      \\ 
V$\text{P}_{25}$T               & 76.8                     &95.5               & 66.6                     &86.1                     &95.9            & 84.2     \\ 
V$\text{P}_{50}$T               & 77.0                     &\textbf{96.0}               & 66.4                     &84.0                     &95.8            & 83.8     \\
\bottomrule
\end{tabular}
}
\vspace{-.3em}
\end{table}


\subsection{Training strategy}
\label{sec:training}
In this section, we investigate the impact of various ``old tricks" in transferable adversarial learning on the performance of visual prompting. We first examine the effect of augmentation methods such as RandomHorizontalFlip, RandAug, and Cutmix on the CIFAR100 dataset. As shown in \cref{table:aug}, interestingly, we find that using simple augmentation techniques like RandomHorizontalFlip can achieve satisfactory results, while more advanced methods such as Cutmix or RandAug may decrease performance, likely due to over-regularization. For example, on CIFAR100, RandomHorizontalFlip is able to improve accuracy by 0.7\%, but RandAug or CutMix hurts the accuracy by 1.1\% and 0.8\%, respectively.

Next, we conducted an ablation study on different gradient normalization strategies, including the $L_1$ norm, $L_2$ norm, and $L_\infty$ norm, on the CIFAR100 dataset. The results, shown in \cref{table:norm}, indicate that the $L_2$ norm consistently performs best among the three strategies. Furthermore, we investigated the optimal configuration for the $L_2$ norm, and found that using the whole gradient of the image to calculate the norm ($L_2$-whole) performs better than using only the gradient of the visual prompting pixels ($L_2$-partial), \ie, 81.2\% \vs 79.4\%.

\begin{table}[]
\caption{\textbf{Ablation on augmentation.} We use CLIP-Base/32 as pre-trained model and evaluate on CIFAR100. We find that simple techniques like RandomFlip achieve strong results, while stronger augmentations like Cutmix or RandAug decrease the performance.}
\vspace{0.2em}
\label{table:aug}
\centering
\resizebox{0.8\linewidth}{!}{
\begin{tabular}{ccc|c}
\hline
\multicolumn{3}{c|}{Augmentation} & \multirow{2}{*}{Performance} \\ \cline{1-3}
Flip  & RandAug   &    CutMix &                          \\ \hline
 \textcolor{light-gray}{\XSolidBrush}          &     \textcolor{light-gray}{\XSolidBrush}   &   \textcolor{light-gray}{\XSolidBrush}    & 80.5                         \\ \hline
\CheckmarkBold        &   \textcolor{light-gray}{\XSolidBrush}     &   \textcolor{light-gray}{\XSolidBrush}   & \textbf{81.2}                         \\ \hline

\CheckmarkBold          &      \CheckmarkBold     &  \textcolor{light-gray}{\XSolidBrush}  & 79.4                         \\ \hline
\CheckmarkBold         &      \textcolor{light-gray}{\XSolidBrush}     &  \CheckmarkBold  & 79.7                         \\ \bottomrule

\end{tabular}}

\end{table}



\begin{table}[]

\caption{\textbf{Ablation on gradient normalization}. We note 1) applying L2 norm on gradient enhance performance; and 2) using the whole image's gradient leads to further improvement compared to using only the gradient of the visual prompting pixels.}
\vspace{0.2em}
\label{table:norm}
\centering
\resizebox{0.88\linewidth}{!}{
\begin{tabular}{cccc|c}
\hline
\multicolumn{4}{c|}{Gradient Normalization} & \multirow{2}{*}{Performance} \\ \cline{1-4}
$L_{1}$  & $L_{\infty}$ & $L_{2}$-partial  & $L_{2}$-whole  &                              \\ \hline
\textcolor{light-gray}{\XSolidBrush}  & \textcolor{light-gray}{\XSolidBrush}     &   \textcolor{light-gray}{\XSolidBrush}      &     \textcolor{light-gray}{\XSolidBrush}   & 77.5   \\ \hline
\CheckmarkBold    &   \textcolor{light-gray}{\XSolidBrush}    &   \textcolor{light-gray}{\XSolidBrush}        &    \textcolor{light-gray}{\XSolidBrush}     & 77.2                         \\ \hline
\textcolor{light-gray}{\XSolidBrush}    &   \CheckmarkBold    &   \textcolor{light-gray}{\XSolidBrush}         &     \textcolor{light-gray}{\XSolidBrush}    & 71.9                         \\ \hline
 \textcolor{light-gray}{\XSolidBrush}   &   \textcolor{light-gray}{\XSolidBrush}    &     \CheckmarkBold       &   \textcolor{light-gray}{\XSolidBrush}      & 79.4                         \\ \hline
 \textcolor{light-gray}{\XSolidBrush}   &   \textcolor{light-gray}{\XSolidBrush}    &       \textcolor{light-gray}{\XSolidBrush}     &  \CheckmarkBold     & \textbf{81.2}                         \\ \bottomrule
\end{tabular}}
\vspace{-1em}
\end{table}

\subsection{VPT-DEEP}

VPT-DEEP is an advanced version of VPT, which additionally introduces learnable tokens at every Transformer layer’s input space for enhancing performance. We hereby briefly compare its performance to that of EVP. Specifically, 
we follow the setup detailed in \cref{subsec: clip} to experiment with the CLIP-base model. Our results indicate that, while VPT-DEEP significantly outperforms the vanilla VPT by 4.6\% (from 77.3\% to 81.9\%), its performance remains slightly lower than that of our EVP, which achieves an average accuracy of 82.5\%. For more detailed accuracy scores across each dataset, please refer to the supplementary material. These findings confirm the efficacy of EVP as a superior prompting strategy. Additionally, given that EVP and VPT-DEEP focus on different aspects of prompting (pixel-level and feature-level, respectively), combining the two methods could lead to the development of even stronger visual prompting techniques. We leave such possibilities for future research.





 


\section{Conclusion}
We propose EVP, a simple and effective method for adapting pre-trained models to various downstream tasks using visual prompts at the pixel level. EVP preserves the original image information and incorporate adversarial learning techniques to improve performance. Our experiments demonstrate that EVP outperforms other visual prompting methods and outperforms linear probing in a variety of settings. Moreover, EVP shows strong performance when dealing with different data scales and robustness to out-of-distribution samples.

{\small
\bibliographystyle{ieee_fullname}
\bibliography{egbib}
}

\end{document}